\documentclass[letterpaper]{article} 
\usepackage{aaai24}  
\usepackage{times}  
\usepackage{helvet}  
\usepackage{courier}  
\usepackage[hyphens]{url}  
\usepackage{graphicx} 
\usepackage{natbib}  
\usepackage{caption} 
\usepackage{algorithm}
\usepackage{algorithmic}
\usepackage{booktabs}
\usepackage{multirow}
\usepackage{amsmath}
\usepackage{rotating}
\usepackage{adjustbox}
\usepackage{amssymb}

\usepackage{newfloat}
\usepackage{listings}
\DeclareCaptionStyle{ruled}{labelfont=normalfont,labelsep=colon,strut=off} 
\floatstyle{ruled}
\newfloat{listing}{tb}{lst}{}
\floatname{listing}{Listing}
\title{Multi-level Multiple Instance Learning with Transformer for Whole Slide Image Classification}
\author{%
Ruijie Zhang\(^{1}\) \quad Qiaozhe Zhang\(^{1}\) \quad Yingzhuang Liu\(^1\) \quad Hao Xin\(^2\) \\
\quad \textbf{Yan Liu}\(^2\) \quad \textbf{Xinggang Wang}\(^{1}\)\thanks{Xinggang Wang is the corresponding author}\\
}
\affiliations{
    \textsuperscript{\rm 1}Huazhong University of Science and Technology \quad \textsuperscript{\rm 2}Ant Group \\

    \texttt{\{k1seki, qiaozhezhang, xgwang\}@hust.edu.cn}\\
}

\usepackage{bibentry}

\begin{document}

\maketitle
\begin{abstract}
Whole slide image (WSI) refers to a type of high-resolution scanned tissue image, which is extensively employed in computer-assisted diagnosis (CAD). The extremely high resolution and limited availability of region-level annotations make employing deep learning methods for WSI-based digital diagnosis challenging. Recently integrating multiple instance learning (MIL) and Transformer for WSI analysis shows very promising results. 
However, designing effective Transformers for this weakly-supervised high-resolution image analysis is an underexplored yet important problem.
In this paper, we propose a Multi-level MIL (MMIL) scheme by introducing a hierarchical structure to MIL, which enables efficient handling of MIL tasks involving a large number of instances. Based on MMIL, we instantiated MMIL-Transformer, an efficient Transformer model with windowed exact self-attention for large-scale MIL tasks. To validate its effectiveness, we conducted a set of experiments on WSI classification tasks, where MMIL-Transformer demonstrate superior performance compared to existing state-of-the-art methods, i.e.,  96.80\% test AUC and 97.67\% test accuracy on the CAMELYON16 dataset, 99.04\% test AUC and 94.37\% test accuracy on the TCGA-NSCLC dataset, respectively. All code and pre-trained models are available at: 
\url{https://github.com/hustvl/MMIL-Transformer}
\end{abstract}

\section{Introduction}

Deep learning's success \cite{hu2019local, huang2019ccnet,ramachandran2019stand,wang2018non,zhao2020exploring,tan2019efficientnet} in sequential image tasks offers promising methods for digital pathology. Whole slide images (WSIs),  which are high-resolution digital slides of tissue specimens, are commonly used in image-based digital pathology \cite{he2012histology}. However, the huge size and absence of pixel/region-level annotations \cite{srinidhi2021deep} in WSIs create obstacles for deep learning-based image diagnosis. To address the aforementioned problem, image-based computer-assisted diagnostics typically employ multiple instance learning (MIL) \cite{keeler1990integrated,dietterich1997solving}, a subset of weakly supervised learning methods \cite{zhou2018brief}.

The dataset of a MIL task is organized into multiple bags, with each bag consisting of multiple instances. The annotations are only available at the bag level. This data organization offers significant benefits in the field of image-based digital pathology. For instance, in cancer diagnosis, the entire WSI (bag-level data) can be labeled as tumor without the necessity of distinguishing normal tissue (instances) within the WSI. This approach substantially alleviates the challenges associated with data and annotations acquisition. However, challenges persist due to the high resolution of WSI and the unbalanced distribution of instance categories.

\begin{figure}[htp]
    \centering
    \includegraphics{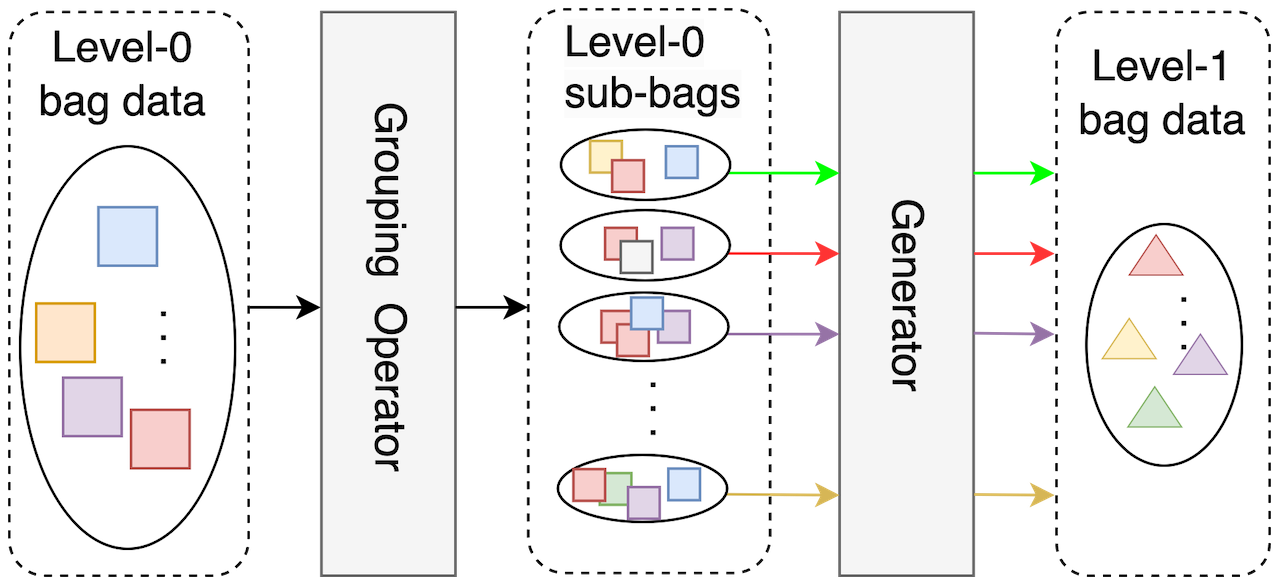}
    \caption{A two-level Multiple Instance Learning framework. 1) A bag is divided into several sub-bags by a grouping operator; 2) Perform multi-head self-attention (MSA) within sub-bags; 3) Making use of sub-bags in 2) to build higher-level bag data;  4) Hierarchical bags can be generated by repeating 1), 2) and 3).}
    \label{fig:hierarchical MIL}  
\end{figure}

Previous studies have explored the applications of MIL in digital pathology tasks. \cite{wang2018revisiting,kanavati2020weakly} perform pooling operation on instance feature embeddings then to following tasks. \cite{ilse2018attention} introduce the attention mechanism, where trainable attention weights are given to each instance for aggregation. Li et al. \cite{li2021dual} introduced non-local attention to MIL, where the similarity is measured between the instance to give distinct attention weights. To introduce correlated information between instances, \cite{shao2021transmil} uses Nyström Attention \cite{xiong2021nystromformer} based transformer to calculate global attention between all instances in one bag.

Transformer \cite{vaswani2017attention} adopts the self-attention mechanism, which calculates the pairwise correlation between each token within a sequence. Self-attention mechanism enables Transformer to effectively capture both the spatial morphology features of individual patches, as well as the correlation information between different patches \cite{chen2021transunet}. However, when working with high-resolution WSIs, Transformer would yield a prohibitively large number of patches, making it challenging to use the original self-attention mechanism. To address this problem, one approach is to use approximate attention \cite{xiong2021nystromformer, beltagy2020longformer}, but may not capture all the intricate relationships within the sequence compared to the original self-attention mechanism. Another approach is to perform local-attention \cite{jaderberg2015spatial,liu2021swin}, which can be classified into two strategies. The first strategy is a non-overlapped approach, reducing computational complexity to a great extent but lacking the capability to capture global information. The second strategy is overlapped approach, which enhances communication among local elements. Nonetheless, as the sequence length grows, this strategy introduces supplementary computational complexity \cite{fang2022msg}.

In response to the aforementioned challenges, we propose the Multi-level MIL (MMIL), aiming at enabling non-approximate self-attention as well as building accurate local-to-global self-attention for large-scale multi-instance learning, i.e., the size of input instances can be up to more than 50k. Based on this strategy, we instantilized MMIL-Transformer for WSIs classification. We conducted a set of experiments on two public datasets, i.e., CAMELYON16 \cite{bejnordi2017diagnostic} and TCGA-NSCLC \cite{tomczak2015review} to demonstrate its effectiveness. The proposed model achieved a promising test AUC 96.80\% and test accuracy 97.67\% on the CAMELYON16 dataset, test AUC 99.04\% and test accuracy 94.37\% on the TCGA-NSCLC dataset. We expect our efforts can further ease the research and application of Transformers for large-scale MIL. 
We summarize our contributions as follows:
\begin{itemize}
    \item We propose a differentiable multi-level MIL (MMIL) formulation, in which original instances can be flexibly grouped into sub-bags and each sub-bag will generate a new instance for final bag classification. Besides, various instance grouping methods, like \textit{MSA grouping}, are proposed and studied.
    \item We propose MMIL-Transformer to conquer the high-resolution WSI classification problem, in which exact self-attention can be computed within local regions and the global WSI representation can be learned effectively. In addition, we design a masking mechanism by removing original instances to further boost performance and reduce computational costs.
    \item The proposed MMIL Transformer can obtain state-of-the-art WSI classification results on the challenging CAMELYON16 and TCGA-NSCLC datasets using both ResNet and ViT as patch encoders.
\end{itemize}


\section{Related Work}

\label{related work}

\subsection{Whole Slide Images Pre-processing}
WSI scanners are capable of converting biopsy slide tissue into a gigapixel image by capturing the image in small, overlapping sections and stitching them together to form a high-resolution image of the entire slide, typically at 20\(\times\) to 40\(\times\) magnification. These images fully preserve the original tissue structure, which is crucial for computer-assisted diagnostics (CAD) \cite{he2012histology}. To read or manipulate WSI, there are several open-source tools available, such as OpenSlides \cite{goode2013openslide} and QuPath \cite{bankhead2017qupath}. However, due to the presence of a large number of background images and noise images in WSI, it is necessary to perform corresponding processes on them. There are two main categories of methods for WSI processing. The first one is a classic image processing method, where the clipped WSI patches are converted to the HSV color space, and then subjected to background removal and denoising based on hue, Saturation and value \cite{tian2019computer}. The second category is a deep learning based method \cite{lu2021data, li2021dual}, where a classifier like CNN is trained to distinguish whether the clipped patches are tissue or not.

\subsection{Multiple Instance Learning in Pathology}
In the pathology field, MIL methods can be categorized into two types, instance-based MIL \cite{kanavati2020weakly, campanella2019clinical, xu2019camel,lerousseau2020weakly,chikontwe2020multiple} and embedding-based MIL \cite{lu2021data,shao2021transmil,li2021dual}, based on the input data for aggregation module.
Instance-based MIL methods first map extracted patches to pseudo-labels that correspond to the bag-level annotation, and then use these corresponding top-k instances for aggregation. Due to the unbalanced distribution of different instance types, instance-based methods always require a large number of WSIs. 
Embedding-based MIL methods first map extracted patches to embeddings then fed all patch feature embeddings to the following modules. Recently, attention mechanism have gained interest in MIL. \cite{ilse2018attention,tomita2019attention,hashimoto2020multi,naik2020deep,lu2021data} introduce trainable weights to each instance. To further improve the performance, non-local attention \cite{li2021dual} and self-attention \cite{shao2021transmil} is also adopted in MIL. Most previous MIL methods focused on evaluating the labels of instances, \cite{wang2018revisiting} emphasizes the representation of bags. DGMIL\cite{qu2022dgmil} considers the feature distribution of WSIs and uses this information to guide MIL. DTFD-MIL\cite{zhang2022dtfd} introduces the concept of pseudo-bags to enlarge the number of bags. \cite{lin2023interventional} addresses MIL from a novel perspective via analyzing the confounders between bags and labels.

\subsection{Self-attention Mechanism}
The inception of Transformer \cite{vaswani2017attention} can be traced back to the domain of natural language processing (NLP), where the self-attention mechanism plays a pivotal role in establishing relations between local features. 
For gigapixel images, the patch sequence becomes too long to process in 
self-attention mechanism. Local-attention and approximate-attention can be employed to address this problem. \cite{beltagy2020longformer}introduce a fixed-size sliding window to sequences, MSA is only performed within the windows. \cite{child2019generating} propose sparse-Transformer, where a subset of the input sequence is selected by adaptive mechanisms to calculate self-attention. \cite{wang2020linformer} introduce a low-rank factorization of the attention matrix. There are also other local-attention methods \cite{kitaev2020reformer,zaheer2020big}. However, local-attention methods present a trade-off dilemma where non-overlapped windows lack communication between each window, while overlapped windows make more computational burden as the sequence goes longer. And the adoption of approximate attention in models results in information loss, reducing the interpretability of the model. Therefore, we require a more efficient, simple, and flexible approach to building local-global relations to handle high-resolution images like WSIs. \cite{fang2022msg} proposed a messenger mechanism to the non-overlapped window attention method, which enables the information exchange between separated windows. 

In MMIL-Transformer, we incorporate \(MSG\) mechanism\cite{fang2022msg} into MMIL as one method to establish a hierarchical framework. However, We do not treat \(MSG\) tokens only as a hub for communication between fixed windows. We use those tokens to generate higher-level instances, which finally contribute to higher-level bags. Moreover, we introduce an embedding-wise masking mechanism to further reduce the number of instances and boost the performance.
\section{Method}
\label{Method}

\begin{figure*}[t]
    \centering
    \includegraphics{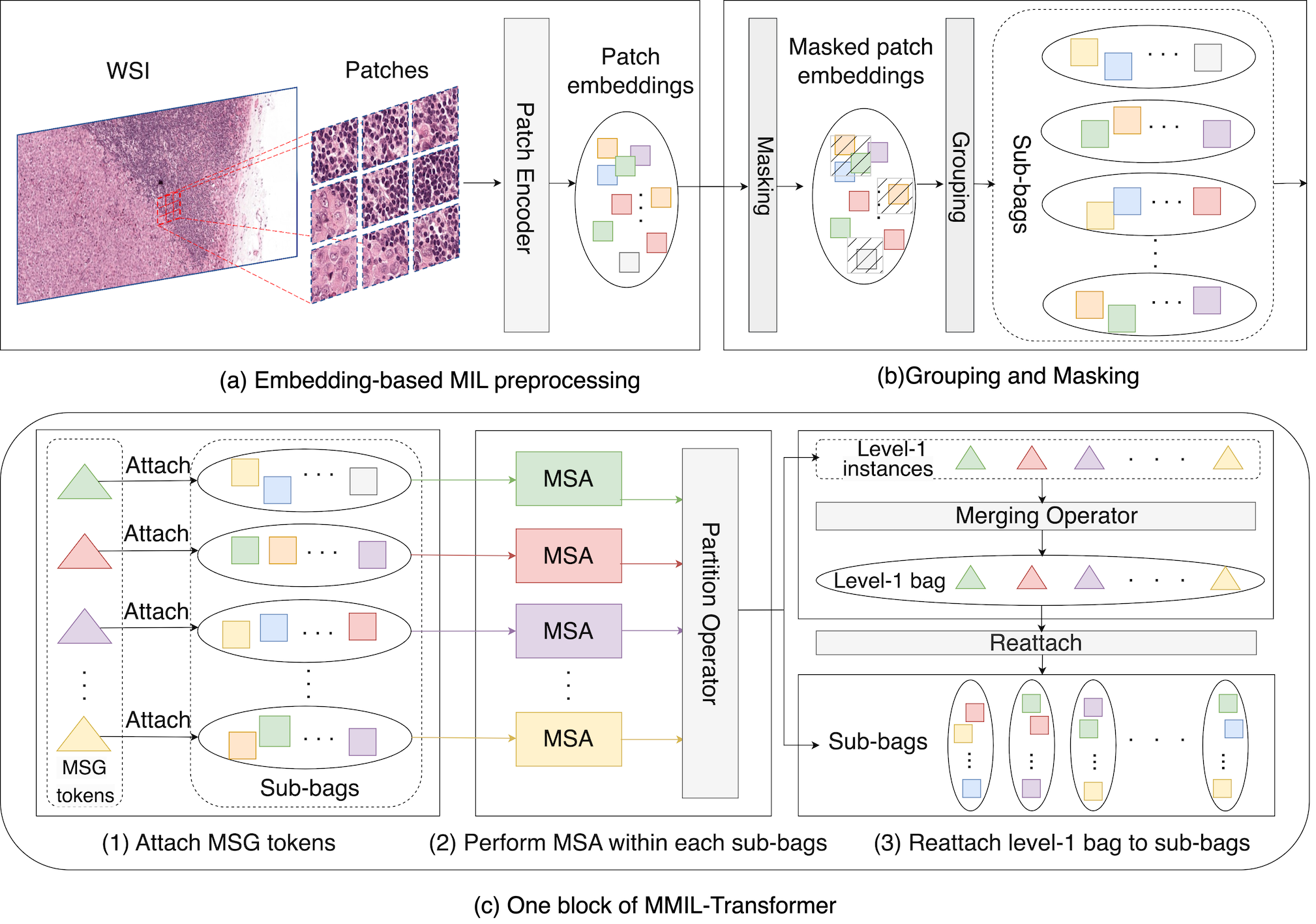}
    \caption{Overview of a two-level MMIL-Transformer. (a) WSI Pre-processing: 1) Background removal; 2) Clip tissue images into patches; 3) Mapping patches to feature embeddings. (b) Grouping and Masking: split feature embeddings into sub-bags and mask fixed ratio of embeddings; (c) One block of MMIL-Transformer: (1) Attach \(MSG\) tokens to each sub-bag. (2) Perform MSA within each sub-bag then part sub-bags from \(MSG\) tokens; 3) Merge Parted \(MSG\) tokens to build a level-1 bag, then reattach all tokens in level-1 bag to level-0 sub-bags;}
    \label{fig:Overview}
\end{figure*}

\subsection{Transformer-based Multiple Instance Learning}

\paragraph{MIL Problem Formulation} 
In MIL, given \(N\) bag-level samples \(\{\mathbf{X}_1,\mathbf{X}_2,\ldots,\mathbf{X}_N\}\) and their corresponding bag-level label\(\{\mathbf{Y}_1,\mathbf{Y}_2,\ldots,\mathbf{Y}_N\}\), each bag-level data \(\mathbf{X}_i\) is made of numbers of instances \(\{\mathbf{x}_{i,1},\mathbf{x}_{i,2},\ldots,\mathbf{x}_{i,n_i}\}\), where the instance-level labels \(\{\mathbf{y}_{i,1},\mathbf{y}_{i,2},\ldots,\mathbf{y}_{i,n_i}\}\)are unknown. Suppose we have a model denoted by \(f(\theta)\), a binary MIL classification task is to give bag-level prediction \(\mathbf{\hat{Y}}_i = f(\mathbf{X}_i;\theta) \in \{0,1\}\) as close to the bag-level label \(\mathbf{Y}_i \in \{0,1\}\ \) as possible, i.e., \(\mathop{\arg\min}_{\theta}L(\mathbf{\hat{Y}}_i,\mathbf{Y}_i)\), where \(L\) denotes the loss function and \(n_i\) is the number of instances of i-st bag-level data. The MIL constraints are illustrated as follows.
\begin{equation}
    {
    \small
    {\bf Y}_i =
        \left\{
            \begin{aligned} 
            x = 0, &\text{ iff }\sum\mathbf{y}_{i,j}=0 \text{,}&\mathbf{y}_{i,j} \in \{0,1\} \text{,} \\
            y = 1, &\text{ otherwise} &
            \end{aligned}
        \right.
    }
\end{equation}

\paragraph{Apply Transformer to MIL-based WSIs Classification}
In the context of a Transformer-based binary classification problem on WSIs, we define bag-level data as referring to WSIs, where label 0 denotes normal WSI and label 1 denotes WSI containing tumors. 
To reduce computational complexity, we can employ an embedding-based approach, where we initially fed segmented patches to a patch encoder extractor (e.g., ResNet\cite{he2016deep}) and use the resulting feature sequences as input to the Transformer. Due to the gigapixel of WSI, the resulting feature sequences are too long to perform MSA. Therefore, we propose a multi-level MIL framework for applying Transformer.

\subsection{Build Multi-level Bags for WSIs}
In this section, we demonstrate how to build hierarchical bags for WSI. The construction process involves two primary stages,  aiming at building a higher-level bag from its corresponding lower-level counterpart. The first stage is grouping and masking, where the lower bag is first divided into sub-bags by a grouping operator with a fixed masking ratio. In the second stage, these sub-bags are used to produce higher-level bags by a generator. We instantiated a messenger-based\cite{fang2022msg} generator, where \(MSG\) tokens are the raw material to produce a higher-level bag.

\paragraph{Grouping Operators}
\label{Grouping operator}
The utilization of grouping operators allows us to partition lengthy sequences into shorter sub-bags and introduce prior information to the data, thus enabling performing MSA. We instantiated types of grouping operators. Each of these operators possesses its own set of pros and cons. 
\begin{itemize}
    \item \textit{Coordinate grouping}: we employ the original coordinates of patches as the basis for grouping. Specifically, we preserve the central coordinates of all clipped patches from a WSI when pre-processing, and use these coordinates for K-Means \cite{hartigan1979algorithm} clustering. Sub-bags are created by the index of clustering results.
    \item  \textit{Embedding grouping}: we use the cosine distance between embeddings for clustering, then treat the result as sub-bags. This approach does not require additional coordinate information.  
    \item \textit{Random grouping}: we randomly group the embeddings to specific numbers of sub-bags.
    \item \textit{Sequential grouping}: the embeddings are grouped into numbers of sub-bags in sequential order, with the original patches arranged on the WSI from left to right and top to bottom.
    \item \textit{MSA grouping}: the embeddings are grouped into numbers of sub-bags according to attention scores produced by multi-head self-attention (MSA). First, we use \textit{Random grouping} to make a fixed number of sub-bags, then perform MSA within each sub-bag to obtain attention scores. Select \(top-k\) instances within each sub-bag, the selected instances from one original group are regarded as a new group, which is presented as Alg. ~\ref{algo: MSA grouping}. This enables grouping operators can be supervised.

\begin{algorithm}[tb]
\caption{\textit{MSA Grouping}}
\label{algo: MSA grouping}
\textbf{Input}: Bag-level data \(\mathbf{X}\) \\
\textbf{Output}: Sub-bags \(\{{G_1},{G_2},\ldots,{G_g}\}\) \\
\begin{algorithmic}[1] 
\STATE Grouping \(\mathbf{X}^{l}\) to \(g\) sub-bags by \textit{Random grouping} \(r\);\\
            \quad\(\{{G_1},{G_2},\ldots,{G_g}\} \leftarrow r(\mathbf{X_h}^{l})\)\\
\WHILE{\(j\) \textless  \(g\)}        
\STATE Perform MSA within each sub-bag \(G_j\) to obtain attention scores \(S_j\);\\
           \quad \(S_j\)  \(\leftarrow\) MSA(\(G_j\))\\
\STATE According to \(S_j\), select \(top-k\) instances from \(G_j\) to build a new group;\\
           \quad \(H_j \leftarrow Top_k(G_j)\) \\
\STATE \(G_j\) is divided to \(H_j\) and \(L_j\), where \(G_j\) = \(H_j\) + \(L_j\);\\
           \quad \(G_j \leftarrow (H_j,L_j) \) \\
\ENDWHILE
\end{algorithmic}
\end{algorithm}
\end{itemize}
\textit{Coordinate grouping} introduces additional information into sub-bags. \textit{Embedding grouping} considers the similarity between patches and does not require an additional step in pre-processing. But these clustering-based grouping methods incur higher computational resource costs. \textit{Random grouping} is simple, and it may bring a benefit for unbalanced data problems. But the randomness of this method sometimes makes the performance unstable when training. \textit{MSA grouping} requires additional computational costs, however, we can supervise grouping by applying specific prior bias. Additionally, grouping operators can be designed freely for different requirements.
\paragraph{Embedding-wise masking}
Corresponding with grouping operators, we further introduce a \textit{masking mechanism} to speed up the model and enhance its performance. Specifically, after partitioning sequences into sub-bags, a fixed ratio of the sub-bags will be masked. These masked sub-bags will be deposed for the following processing.

\paragraph{Build a higher-level instance}
There are many types of methods that can be used to build higher-level instances. For example, downsampling and pooling. We instantiated one higher-level instance generator by using the messenger mechanism \cite{fang2022msg}. As shown in Fig.~\ref{fig:Overview}(c), \(m\) \(MSG\) tokens \({{T}_{MSG}^j} \in \mathbb{R}^{m{\times}C}\) will be attached to each sub-bag \(G_j \in \mathbb{R}^{w_j{\times}C}\) as \({G_j}^{\prime}\in{\mathbb{R}^{(w_j+m)\times{C}}}\), where \(C\) denotes the embedding channel of patch feature, \(w_j\) denotes the number of members in \(j\)-th sub-bag. Therefore, one WSI is divided into \(MSG\) token attached sub-bags \(\{{G_1}^{\prime},{G_2}^{\prime},\ldots,{G_g}^{\prime}\}\). MSA is performed within each sub-bag between both feature embeddings and \(MSG\) tokens. \(MSG\) tokens can capture information from the corresponding sub-bag with attention. Finally, all \(MSG\) tokens from different sub-bags are collected as higher-level instances, then undergo a merging operation to produce a higher-level bag. Moreover, a class (\(CLS\)) token is attached to the higher-level bag for the classification task. Actually, this \(CLS\) token can be regarded as the final-level bag.

\paragraph{Merging Operation}
Merging operation aims to collect and merge instances generated from sub-bags for producing a higher-level bag. There are many merge operations, including but not limited to shuffle and average. Corresponding with the \(MSG\) based generation method, we implement self-attention to merge instances. In this case, after performing MSA with patch embeddings, the patch embeddings are partitioned from the \(MSG\) tokens, and the \(MSG\) tokens are subsequently fed into a Transformer to build higher-level bags.

Overall, the process to generate multi-level bags for WSI can be summarized as Alg.~\ref{algo: multi-level bags}.
\begin{algorithm}[tb]
\caption{Bulid \(k\)-level bags though \(MSG\) tokens}
\label{algo: multi-level bags}
\textbf{Input}: Bag-level data \(\mathbf{X_h}^{0}\)\\
\textbf{Output}: Set of \(k\) level bags \(\mathbf{{X_h}}^{K}\)
\begin{algorithmic}[1] 
\WHILE{level \textless  \(k\)}
\STATE Grouping and masking \(\mathbf{X_h}^{l}\) to \(g\) sub-bags by operation \(c\);\\
            \quad\(\{{G_1},{G_2},\ldots,{G_g}\} \leftarrow c(\mathbf{X_h}^{l})\)\\
\STATE  Attach \(MSG\) tokens to each sub-bags;\\
        \quad \({{G_j}^{\prime}} \leftarrow \text{Concat}({{T}^j_{MSG}};G_j),\quad j \in \{1,2,\ldots,g\}\);
\STATE Perform MSA in each sub-bag;\\
        \quad \({{G_j}^{\prime}} \leftarrow \text{MSA}({G_j}^{\prime}) + {G_j}^{\prime}\);
\STATE Merging all \(MSG\) tokens to build higher level bag;\\
           \quad \({\mathbf{{X_h}}^{l+1}} \leftarrow \text{Merge}(T^1_{MSG},T^2_{MSG},\ldots,T^g_{MSG})\)

\STATE Append  \({\mathbf{{X_h}}^{l+1}}\) to \(\mathbf{{X_h}}^{K}\).
\ENDWHILE
\STATE Attach \(CLS\) token to highest level bag;\\
        \quad \(\mathbf{X_h}^{k}\leftarrow\) Concat(\(\mathbf{X_h}^{k},T_{cls}\))

\STATE \textbf{return} \(\mathbf{X_h}^{K}\)
\end{algorithmic}
\end{algorithm}
\subsection{MMIL-Transformer}

We now present MMIL-Transformer. As shown in Fig.~\ref{fig:Overview}, we first preprocess WSI following the procedure mentioned in Fig.~\ref{fig:Overview}(a), which converts high-resolution WSI to patch embeddings. These embeddings are fed into MMIL-Transformer layers. In one basic layer, higher-level bags are generated through Alg.~\ref{algo: multi-level bags}. It is worth noting that the higher-level bags are derived from the \(MSG\) tokens obtained from lower-level sub-bags. This relationship allows for their reattachment to the corresponding lower-level sub-bags, thereby enabling the implementation of a multi-layer design for MMIL-Transformer. Finally, the model will give a prediction to one WSI by performing MLP to the class token \({T_{cls}}\).
We summarize the whole MMIL-Transformer as Alg.~\ref{algo:MMIL}.

\paragraph{Complexity Analysis} 
For a sequence \(S \in \mathbb{R}^{p \times C} \), where \(p\) is the length of \(S\), \(C\) is number of channels. The complexity of the original MSA per layer is \(\mathcal{O}(p^2 \times C)\). Assuming we divide \(S\) into \(g\) sub-bags with a masking ratio of \(\ 0 \leq r < 1\), denoted as \(G \in{\mathbb{R}^{g\times\frac{p}{g}\times (1-r) \times C}}\). The complexity of MSA per sub-bag is \(\mathcal{O}((\frac{p(1-r)}{g})^2 \times C)\). Thus, for one specific level bag, the total complexity of MSA per layer in MMIL-Transformer is \(\mathcal{O}(\frac{1}{g}\times {p^2(1-r)}^2 \times C)\).

\begin{algorithm}[tb]
\caption{MMIL-Transformer}
\label{algo:MMIL}
\textbf{Input}: Bag of instances \(\mathbf{X_i}=\{x_{i,1},x_{i,2},\ldots,x_{i,n}\}\)\\
\textbf{Output}: Bag-level prediction \(\mathbf{\hat{Y_i}}\)
\begin{algorithmic}[1] 
\STATE Embed the instances to feature space by \(h\);\\
           \quad \(\mathbf{X_h} \leftarrow h(\mathbf{X_i})\), where \(\mathbf{X_h} \in \mathbb{R}^{n \times C}\);
\FOR{each layer} 
\STATE Build \(k\)-level bags \(\mathbf{X_h}^{K}\) by Alg.~\ref{algo: multi-level bags}, denoted by MLB;\\
           \quad\({X_h^{K}}=\text{MLB}(X_h)\), where \({X_h^{K}} = \{ {X_h^{0}},\ldots,{X_h^{k}} \}\);
\STATE Reattach \(X_h^{l+1}\) to \(X_h^{l}\) for all \(X_h^{l} \in X_i^{K}\);\\
          \quad \(X_h^{l} = \text{Concat}(X_h^{l+1},X_h^{l})\), for \(l \in\{0,1,\ldots,k\}\); 
\STATE Fed \(X_h^{0}\) to next layer;
\ENDFOR
\STATE Give prediction \(\hat{\mathbf{Y}}\) though \(T_{cls}\);\\
            \quad\(\hat{\mathbf{Y}} \leftarrow \text{MLP}(T_{cls})\)
\STATE \textbf{return} \(\hat{\mathbf{Y}}\)
\end{algorithmic}
\end{algorithm}

\section{Experiments}
\label{Experiments}
\begin{table*}[htb]
  \caption{Results on CAMELYON16 and TCGA-NSCLC using ResNet-50 as the patch encoder}
  \label{table:performance_comparison}
  \centering
  \begin{adjustbox}{width=0.7\linewidth}
  \begin{tabular}{cccccc}
    \toprule
    
    \multirow{2}{*}{Method} & \multirow{2}{*}{Publication} & \multicolumn{2}{c}{CAMELYON16} & \multicolumn{2}{c}{TCGA-NSCLC}       \\
    \cmidrule(r){3-6}
    & & Accuracy & AUC & Accuracy & AUC\\
    \midrule
    ABMIL~\cite{ilse2018attention} &ICML 2018 & 0.8682 & 0.8760 & 0.7719 & 0.8656\\
    PT-MTA~\cite{li2019patch} &MICCAI 2019& 0.8217 & 0.8454 & 0.7379 & 0.8299\\
    MIL-RNN~\cite{campanella2019clinical} & Nature Med 2019& 0.8450 & 0.8880 & 0.8619 & 0.9107\\
    DSMIL~\cite{li2021dual} &CVPR 2021 &0.7985 & 0.8179 & 0.8058 & 0.8925\\
    CLAM-SB~\cite{lu2021data}  &Nature BME 2021 & 0.8760 & 0.8809 & 0.8180 & 0.8818\\
    CLAM-MB~\cite{lu2021data}  &Nature BME 2021 & 0.8372 & 0.8679 & 0.8422 & 0.9377\\
    TransMIL~\cite{shao2021transmil} &NeurIPS 2021 & 0.8837 & 0.9309 & 0.8835 & 0.9603\\
    DGMIL~\cite{qu2022dgmil} &MICCAI 2022 & 0.8018 & 0.8368 & 0.9200 & 0.9702\\
    DTFD-MIL (AFS) \cite{zhang2022dtfd}  &CVPR 2022 &0.908 & 0.946 & 0.891 & 0.951\\
    DTFD-MIL (MaxS) \cite{zhang2022dtfd}  &CVPR 2022 & 0.899 & 0.941 & 0.894 & 0.961\\
    MSG-Transformer \cite{fang2022msg} &CVPR 2022 & 0.8062 & 0.8313 & 0.9354& 0.9874\\
    {MMIL-Transformer} & - & \textbf{0.9341} & \textbf{0.9474} & \textbf{0.9437} & \textbf{0.9904}\\
    \bottomrule
  \end{tabular}
  \end{adjustbox}
\end{table*}
To demonstrate the superior performance of the proposed MMIL-Transformer, we conducted various experiments on the CAMELYON16 \cite{bejnordi2017diagnostic} dataset and TCGA-NSCLC \cite{tomczak2015review} dataset.

\paragraph{Dataset and Evaluation Metrics}
CAMELYON16 is a public dataset for metastasis detection in hematoxylin and eosin (H\&E) stained WSIs of lymph node sections. The dataset contains 270 training WSIs and 129 test WSIs. We perform pre-processing at \(\times\)20 magnification, which produces 4.6 million \(256 \times 256\) non-overlapping patches. All patches are mapped to feature embeddings by ImageNet pre-trained ResNet50 \cite{he2016deep} baseline.

The TCGA-NSCLC dataset focuses on non-small cell lung cancer (NSCLC), encompassing two types of carcinoma: Lung Squamous Cell Carcinoma (TGCA-LUSC) and Lung Adenocarcinoma (TCGA-LUAD). TCGA-NSCLC comprises a total of 1053 diagnostic WSIs, with 541 slides from 478 cases for LUAD and 512 slides from 478 cases for LUSC. For pre-processing, we adopt the same configuration as DSMIL \cite{li2021dual}, where a total of 1046 WSIs are involved in the computation. We split the dataset in the ratio of training:validation:test = 60:15:25, resulting in 627 training WSIs, 156 validation WSIs and 263 test WSIs.

To evaluate the classification performance, we utilized accuracy and area under the curve (AUC) scores as the evaluation metrics. In all experiments, the accuracy was calculated using a fixed threshold of 0.5.

\paragraph{Experiments Details}
We employ cross-entropy as our loss function, and the Adma optimizer was employed with a learning rate of 1e-4 and weight decay of 1e-5. The default number of basic layers is 2. For the CAMELYON16 dataset, we grouped the embeddings into 10 sub-bags using the \textit{random grouping} operator and randomly masked the sub-bags by a ratio of 0.6. As for the TCGA-NSCLC dataset, we employed \textit{random grouping} to split the embeddings into 4 sub-bags. In both datasets, one single \(MSG\) token is attached to each sub-bag. These \(MSG\) tokens are then used to construct higher-level bags structure for later modules. All experiments were conducted using a 24\(GB\) RTX 3090 GPU.

\subsection{Results on WSI Classification}
The binary classification tasks encompass the identification of positive and negative cases within the CAMELYON16 dataset, LUSC and LUAD subtypes within the TCGA-NSCLC dataset. All results are presented in Tab.~\ref{table:performance_comparison}. Notably, except MMIL-Transformer(CTransPath) uses CTransPath as the patch encoder, all others use ResNet50. Compared with other state-of-the-art methods wo \cite{shao2021transmil,lu2021data,li2021dual,campanella2019clinical,ilse2018attention,li2019patch,zhang2022dtfd,qu2022dgmil}, the proposed MMIL-Transformer consistently achieves superior results in both accuracy and AUC score on both datasets. Note that all test experiments are conducted 10 times in test datasets to calculate the average accuracy and AUC.

The CAMELYON16 dataset exhibits an imbalanced distribution of normal tissue patches and tumor tissue patches, posing significant challenges for the classification task. However, MMIL-Transformer overcomes these challenges and achieves a notable improvement of 3.51\% accuracy compared to DTFD-MIL(MaxMinS). Compares the approximate attention method TransMIL, MMIL-Transformer achieves an  improvement of 5.04\% accuracy and 1.65\% AUC. CLAM, PT-MTA, and ABMIL do not consider the correlation between instances, which leads to limited performance. In comparison to other non-attention methods, the utilization of clustering operation in CLAM has enabled it to achieve relatively better performance. While MSG-Transformer adeptly employs the \(MSG\) for local information exchange, its performance remains constrained in the absence of accurate local-to-global self-attention.

In the TCGA-NSCLC dataset, there is a higher abundance of positive tissue patches compared to negatives, resulting in relatively easier identification of LUSC with LUAD. All mentioned methods achieve relatively better performance than their performance on CAMELYON16. However, MMIL-Transformer stands out with an impressive accuracy of 94.37\% and AUC score of 99.04\%. 

We also compare MMIL-Transformer with other methods by using CTransPath~\cite{wang2022transformer} as the patch encoder. The results are shown in Tab. 
~\ref{table:featurextractor}. In this part, we use the same dataset configurations as \cite{lin2023interventional}. We use \textit{MSA grouping} with a masking ratio of 0.2 to make 10 sub-bags for MMIL-Transformer on CAMELYON16, \textit{MSA grouping} without masking to make 4 sub-bags on TCGA-NSCLC. Compared with the most recent works like IBMIL(\cite{lin2023interventional}), our methods demonstrate competitive performances. Compared with IBMIL-TransMIL, MMIL-Transformer achieves 97.67\%(+1.55\%) on the CAMELYON16 and 94.28\%(+0.47\%) on the TCGA-NSCLC. However, MMIL-Transformer is slightly lower than IBMIL-TransMIL on the AUC, with a mere -0.2\% on the CAMELYON16.
\begin{table}[h]
  \caption{Results on CAMELYON16 and TCGA-NSCLC using CTransPath as the patch encoder}
  \label{table:featurextractor}
  \centering
  \begin{adjustbox}{width=\columnwidth}
  \begin{tabular}{cccc}
    \toprule
    Dataset & Method & Accuracy & AUC      \\
    \midrule
    \multirow{4}{*}{{CAMELYON16}}
                        & TransMIL (NeurIPS 2021) & 0.9457 & {0.9588}\\
                          & DTFD-MIL (CVPR 2022) &0.9535 & 0.9618\\
                          & IBMIL-TransMIL (CVPR 2023) &0.9612 &  \textbf{0.9700}\\
                          & {MMIL-Transformer} & \textbf{0.9767}& {0.9680}\\
    \hline
    \multirow{4}{*}{TCGA-NSCLC} & ABMIL (ICML2018) &0.9048 &{0.9587} \\
                    & IBMIL-DSMIL (CVPR 2023) &{0.9143} &{0.9751}\\
                    & IBMIL-TransMIL (CVPR 2023) &0.9381 &{0.9724} \\
                    & {MMIL-Transformer} &\textbf{0.9428} & \textbf{0.9788} \\
    \bottomrule
  \end{tabular}
  \end{adjustbox}
\end{table}

\section{Ablation Study}

We further conducted a series of ablation studies using ResNet-50 as the patch encoder to determine the contribution of several proposed modules. 

\subsubsection{Effects of Grouping Operator}
\label{effects:grouping}
To determine the impact of different grouping operators, we performed a series of experiments while keeping the other configurations unchanged. Notably, to avoid the effects of randomness, we employ \textit{coordinate grouping} instead of \textit{random grouping} on CAMELYON16, \textit{embedding grouping} instead of \textit{random grouping} on TCGA-NSCLC. The masking ratio is 0 for this part.

We first study the influence of the number of sub-bags. The results are shown in Tab.~\ref{table:Effects of Groupnum}. In CAMELYON16 the test AUC decays as the number of sub-bags increases. This behavior can be attributed to the trade-off between the number of sub-bags and available memory resources. Larger numbers of sub-bags require less memory and computational cost, but it makes obstacles to effective communication between them. It can be predicted that the accuracy will be lower as the number of sub-bags increases. In TCGA-NSCLC, there are relatively fewer patches(with a minimum of 23 patches) in WSIs, we were able to split the dataset into up to 6 sub-bags using clustering. Compared with CAMELYON16, the performance deterioration in TCGA-NSCLC is relatively minor.
\begin{table}[h]
  \caption{Effects of Different Grouping Numbers.}
  \label{table:Effects of Groupnum}
  \centering
  \begin{adjustbox}{width=\columnwidth}
  \begin{tabular}{cccc}
    \toprule
    Dataset & Sub-bags number & Accuracy & AUC      \\
    \midrule
    \multirow{4}{*}{{CAMELYON16}} & 10 & \textbf{0.8992} & \textbf{0.9500}\\
                          & 16 &0.8682 & 0.9305\\
                          & 32 &0.8992 &  0.9314\\
    \hline
    \multirow{3}{*}{TCGA-NSCLC} & 2 &0.9354 &\textbf{0.9896} \\
                    & 4 &\textbf{0.9392} &0.9887 \\
                    & 6 &0.9316 &0.9890 \\
    \bottomrule
  \end{tabular}
  \end{adjustbox}
\end{table}

We also studied the effect of different grouping methods, the experiment was conducted on CAMELYON16, and the results are shown in Tab.~\ref{table:Effects of group methods}. An interesting observation is that the incorporation of additional coordinate information leads to a performance decline in terms of accuracy when compared to \textit{random grouping}. In our view, using random grouping alleviates the influence of instance imbalance. Through the process of \textit{random grouping}, each sub-bag has the potential to be allocated positive instances. This enables the attention mechanism to focus on the positive instances within each sub-bag, thereby enhancing the accuracy of the model. Conversely, the incorporation of positional information may give rise to a substantial quantity of sub-bags that lack positive instances, thereby exacerbating the issue of imbalance on higher-level bags. On the other hand, embedding grouping, which does not need additional information, yields relatively lower accuracy and AUC scores. When we group the embeddings in sequential order, thereby utilizing ill spatial information, the resulting scores in accuracy and AUC are lower. \textit{MSA grouping} achieves the highest accuracy but requires more computational resources.
\begin{table}[h]
  \caption{Effects of Different Grouping Methods}
  \label{table:Effects of group methods}
  \centering
  \begin{adjustbox}{width=\columnwidth}
  \begin{tabular}{cccc}
    \toprule
    Dataset & Method & Accuracy & AUC      \\
    \midrule
    \multirow{5}{*}{CAMELYON16} & Coordinate grouping & {0.8992} & \textbf{0.9500}\\
                          & Embedding grouping &0.8682 & 0.9216\\
                          & Sequential grouping &0.8604 & 0.9097\\
                          & Random grouping&{0.9178} & {0.9219}\\
                          & MSA grouping&\textbf{0.9224} & {0.9363}\\
                          
    \bottomrule
  \end{tabular}
  \end{adjustbox}
\end{table}
\subsubsection{Effects of Masking}
We conducted several experiments to determine the effects of the masking mechanism. The grouping method is \textit{random grouping} on the dataset. The results are presented in Tab.\ref{table:Effects of masking}. Including the masking mechanism results in a 1.63\% accuracy enhancement in the case of 10 sub-bags number. As the number of sub-bags goes up, accuracy decrease without masking. However, the masking mechanism recoups the performance decline.
\begin{table}[h]
  \caption{Effects of Different Masking Ratio On CAMELYON16}
  \label{table:Effects of masking}
  \centering
  \begin{adjustbox}{width=\columnwidth}
  \begin{tabular}{ccccc}
    \toprule
        &  Sub-bags number & Masking ratio & Accuracy & AUC      \\
    \midrule
     & \multirow{3}{*}{10} & 0.6 & \textbf{0.9341}&\textbf{0.9474}\\
                                                    && 0.3 & 0.9302 & 0.9372 \\
                                                    && 0 &{0.9178} & {0.9219}\\
                                                    \hline
                                & \multirow{3}{*}{16} & 0.6 & \textbf{0.9302}&\textbf{0.9395}\\
                                                    && 0.3 & 0.9147 & 0.9373 \\
                                                    && 0 &{0.8682} & {0.9305}\\
                                                    \hline
                                & \multirow{3}{*}{32} & 0.6 & \textbf{0.9302}&{0.9263}\\
                                                    && 0.3 & 0.9069 & 0.9222 \\
                                                    && 0 &{0.8992} & \textbf{0.9314}\\
 
    \bottomrule
  \end{tabular}
  \end{adjustbox}
\end{table}
\subsubsection{Effects of Multi-level Framework}
In CAMELYON16, patches of different WSIs in \(\times\)20 magnification range from 142 to 55852. In TCGA-NSCLC, patches range from 23 to 14990. The minimum number of patches determines the maximum number of sub-bags that can be formed, consequently limiting the dimension of higher-level bags. Ablation experiments are conducted on CAMELYON16 by using \textit{coordinate grouping} for the reason of its relative more patches. Again, due to the impracticality of directly performing MSA on level-0 bags as the sequences are too long, we conduct this part experiments by using a Nyström Attention method for fairness. The results are shown in Tab.~\ref{table:Effects of Multi-level Framework}. MMIL-Transformer achieves a similar result as TransMIL in 0-level bags due to they both employ approximate attention. The performance on level-1 is much worse than level-0 due to the trade-off mentioned before and the information loss from approximate-attention. The crucial aspect is that this multi-level framework enables the suggested model to perform non-approximate MSA, leading to significantly enhanced performance.
\begin{table}[h]
  \caption{Effects of Multi-level Framework.}
  \label{table:Effects of Multi-level Framework}
  \centering
  \begin{adjustbox}{width=\columnwidth}
  \begin{tabular}{ccccc}
    \toprule
    Dataset & \(k\) & Accuracy & AUC & Approximate-attention    \\
    \midrule
    \multirow{3}{*}{CAMELYON16} & 0 &  0.8759 &0.9227 & \(\checkmark\) \\
                    
                          & 1 &{0.6667} & {0.6650} & \(\checkmark\) \\
                          &1 &\textbf{0.8992} &\textbf{0.9500} & \(\times\) \\
    \bottomrule
  \end{tabular}
  \end{adjustbox}
\end{table}

\section{Discussion and Conclusion}
In our work, we proposed MMIL, a multi-level MIL learning strategy. Based on this strategy, we construct a MMIL-Transformer model for long-sequence MIL tasks. The proposed model first splits bag-level data into sub-bags, then builds hierarchical bags from those sub-bags for the following stages. This enables MMIL-Transformer to apply non-approximate MSA. which significantly enhances its performance. Moreover, the masking mechanism equivalently enhances the data while also further reducing the computational complexity. Overall, MMIL is an effective MIL learning strategy, by which MMIL-Transformer can better establish local-global connections with less complexity. This makes it more suitable for large-scale unbalanced/balanced MIL learning.

\paragraph{Limitations and Future Work} When dealing with CAMELYON16, we find the performance is more sensitive to the grouping results due to the unbalanced problem. There should be a more suitable grouping operator for making the performance more stable. We will explore these issues in the follow-up work.

\bibliography{aaai24}

\end{document}